\documentclass[letterpaper, 10 pt, conference]{ieeeconf}  

\IEEEoverridecommandlockouts                              

\usepackage{cite}
\usepackage{amsmath,amssymb,amsfonts}
\usepackage{times}
\usepackage{algorithm}
\usepackage{algorithmic}
\usepackage{graphicx}
\usepackage{stfloats}
\usepackage{subfigure}

\title{\LARGE \bf
ClusterFusion: Real-time Relative Positioning and \protect\\ Dense Reconstruction for UAV Cluster
}
\author{Yifei Dong$^{1}$, Shuhui Bu$^{1}$*, Kun Li$^{1}$, Lin Chen$^{1}$, Zhenyu Xia$^{1}$, \protect\\ Yu Wang$^{1}$, Pengcheng Han$^{1}$, Xuefeng Cao$^{2}$, Ke Li$^{2}$ 
\thanks{*Corresponding author (bushuhui@nwpu.edu.cn)}
\thanks{$^{1}$School of Aeronautics, Northwestern Polytechnical University}%
\thanks{$^{2}$PLA Strategic Support Force Information Engineering University}%
}

\begin{document}

\maketitle
\thispagestyle{empty}
\pagestyle{empty}

\begin{abstract}

As robotics technology advances, dense point cloud maps are increasingly in demand. However, dense reconstruction using a single unmanned aerial vehicle (UAV) suffers from limitations in flight speed and battery power, resulting in slow reconstruction and low coverage. Cluster UAV systems offer greater flexibility and wider coverage for map building. Existing methods of cluster UAVs face challenges with accurate relative positioning, scale drift, and high-speed dense point cloud map generation. 
To address these issues, we propose a cluster framework for large-scale dense reconstruction and real-time collaborative localization. The front-end of the framework is an improved visual odometry which can effectively handle large-scale scenes. 
Collaborative localization between UAVs is enabled through a two-stage joint optimization algorithm and a relative pose optimization algorithm, effectively achieving accurate relative positioning of UAVs and mitigating scale drift.
%
%
Estimated poses are used to achieve real-time dense reconstruction and fusion of point cloud maps.
To evaluate the performance of our proposed method, we conduct qualitative and quantitative experiments on real-world data. The results demonstrate that our framework can effectively suppress scale drift and generate large-scale dense point cloud maps in real-time, with the reconstruction speed increasing as more UAVs are added to the system. 

\end{abstract}

\section{Introduction}

Real-time large-scale reconstruction using UAVs plays a significant role in various fields, such as urban planning, emergency rescue, disaster relief, and agricultural monitoring. 
%
However, the complex and variable application scenarios require better real-time performance of large-scale 3D dense reconstruction technology while maintaining reconstruction accuracy.
Currently, the fundamental method for reconstruction is structure from motion (SfM), but it often incurs a significant computation and memory cost. In most cases, SfM methods take hours to produce the final results. Moreover, in larger-scale scenarios, the computation time increases dramatically.
Faster dense reconstruction methods based on the simultaneous localization and mapping (SLAM) technique, such as DenseFusion \cite{chen2020densefusion}, rely on precise pose estimation of the UAV at different times. Although real-time dense reconstruction is possible, reconstuction coverage and speed cannot be further improved due to limitations of single UAV flight speed and battery power. UAV cluster can perform mapping tasks more flexibly and efficiently because of their collaborative work mode, which can solve these problems.

\begin{figure}[tb]
	\centering
	\includegraphics[width=1\linewidth]{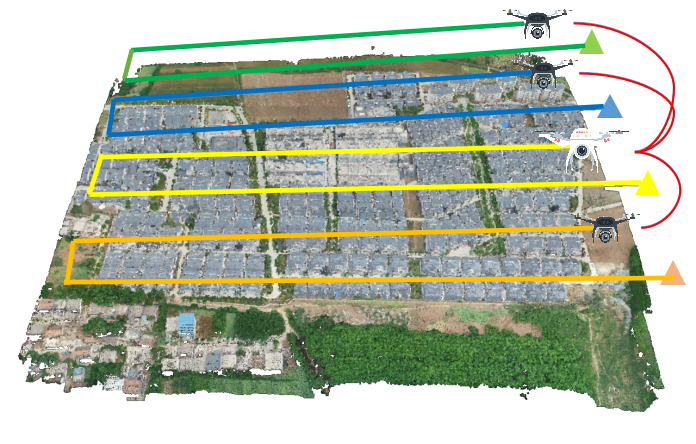}
	\caption{This image shows the real-time dense reconstruction using multiple UAVs. Lines of different colors represent the trajectories of UAVs and the triangles represent the ending points of the trajectories. The white UAV represents the central UAV. The red line represents the relative poses of UAVs.}
	\label{fig:example}
\end{figure}

Real-time point cloud reconstruction on a single UAV is limited by its flight endurance and scanning range. Traditional multi-UAV collaborative SLAM techniques, such as \cite{schmuck2019ccm}, require significant computational resources and communication bandwidth for collaborative pose estimation, which rely on common view relationships.
A good balance between real-time performance and higher accuracy needs to be considered, however, current research seldom addresses this problem due to the following unresolved issues.
For multi-UAV real-time dense reconstruction, the accurate relative poses of multiple UAVs are required, hence the image data collected by multiple UAVs need to be fully utilized for estimating their relative position and for further joint optimization. 
On the other hand, positioning for multiple UAVs has the following challenges: First, to ensure the efficiency of multi-UAV reconstruction, there is often a low overlap of images acquired between multiple UAVs. Second, achieving relative positioning between UAVs using image information alone is difficult until a valid loop-back is detected.

To address the aforementioned challenges, a framework for joint real-time dense reconstruction using multiple UAVs is proposed and developed. For the precise positioning of a single UAV, a large-scale scenes SLAM named as PI-SLAM is developed, which based on GSLAM framework \cite{zhao2019gslam}. Moreover, a hybrid approach is proposed for the relative position optimization of UAVs, which is divided into two parts: distributed optimization and joint optimization.
For UAV lightweight positioning requirement, a two-stage joint GNSS optimization algorithm is applied to a single UAV in the distributed optimization part which can fully utilize visual and GNSS information for initially optimizing the local and global poses. The algorithm clusters UAVs efficiently using GNSS prior information, at minimal computational cost.
In order to obtain more accurate relative poses, they are further optimized in the joint optimization part by using the co-viewing relationships of the images. 
%
%
A demo of dense reconstruction using multiple UAVs can be found in Fig. \ref{fig:example} and the architecture of this framework is shown in Fig. \ref{fig:pipeline}. The contributions of this work are summarized as follows: 
\begin{itemize}
	\item We present a cluster reconstruction framework, which can expand the number of UAVs with low computation and communication costs and perform the collaborative reconstruction tasks in real-time.
	\item We propose a two-stage joint optimization algorithm that can optimize the local pose and scale, as well as estimate the relative poses, without incurring high communication costs.
	\item To achieve more accurate poses estimation between UAVs, a relative pose optimization method is proposed which first detects the common-view relationships between images and then optimizes the relative poses based on homography.
	\item Datasets\footnote{github.com/npupilab/cluster-fusion-dataset} for large-scale 3D dense reconstruction of multiple UAVs in real world are created which is public access.
\end{itemize}

\section{Related Works}

Collaborative pose estimation and dense reconstruction technologies of UAVs involve multiple fields. In this section, we briefly introduce related works.

\subsection{Swarm Localization and Mapping}

In recent years, UAV swarm technology has become a hot research field. There are many methods proposed, especially in visual SLAM field. This section provides an overview of the collaborative visual SLAM framework for multiple robots.

Most approaches to swarm localization and mapping rely on a centralized architecture, which means it is necessary to set up a central node to handle relative pose estimation and mapping. Compared with decentralized architectures investigated in a number of works, centralized architectures are easier to implement and deploy. 
Andersson \cite{andersson2008c} and Kim \textit{et al.} \cite{kim2010multiple} apply a framework that collects all the measurements at a central node and estimates a global map. Based on ORB-SLAM \cite{mur2015orb}, Schmuck \textit{et al.} propose a multi-UAV co-localization framework CCM \cite{schmuck2019ccm}, which successfully solves the associated 3D estimation challenges. CCM maintains a global map and uses the map information to determine the relative pose of a single robot. To limit the memory, Jimenez \textit{et al.} \cite{jimenez2018decentralized} propose a method to distribute the generated maps among multiple robots for storage.

In addition to the centralized architecture, some decentralized cluster SLAM architectures have been proposed. Decentralized SLAM architectures, such as those proposed in \cite{choudhary2017distributed, cieslewski2018data}, offer significant benefits over the centralized architecture in terms of computation, communication, and privacy. Tian \textit{et al.} propose a multi-robot framework Kimera-multi \cite{tian2022kimera}, which achieves estimation errors comparable to those of a centralized SLAM system, while operating in a fully distributed manner.

Existing swarm solutions focus mainly on improving localization performance, with little attention paid to building dense point cloud maps, or at most, building sparse ones. In order to achieve real-time construction of large-scale dense point cloud maps, it is necessary to develop a cluster architecture that integrates both localization and dense mapping capabilities.

\begin{figure}[tb]
	\centering
	\includegraphics[width=1\linewidth]{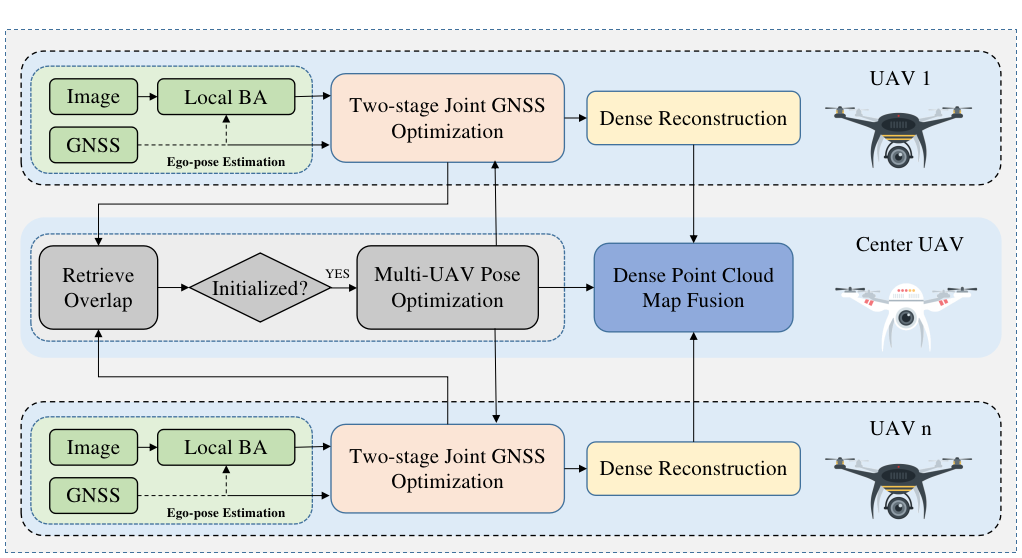}
	\caption{ClusterFusion: The framework for dense reconstruction of multi-UAV. The white UAV represents the center UAV. Each UAV performs ego-pose estimation, global pose optimization and dense reconstruction distributedly, and then multi-UAV joint optimization and point cloud fusion are performed at the central node. }
	\label{fig:pipeline}
\end{figure}

\subsection{Visual SLAM}

There are plenty of visual SLAM approaches proposed. It is impossible to cover all of them in a limited space, therefore, we introduce the most common visual SLAM solutions and the GNSS assisted localization method only. 

The first real-time mono SLAM system is MonoSLAM \cite{davison2007monoslam}, which is based on Extended Kalman Filter (EKF). PTAM \cite{klein2009parallel} is the first SLAM system which parallel the tracking thread and the mapping thread. In addition to tracking and mapping, an additional  looping closing thread is added in ORB-SLAM \cite{mur2015orb}, which can effectively eliminate accumulated error where loop-back is detected. 
There are also some direct method to estimate pose and build semi-dense or dense map, such as \cite{engel2014lsd,engel2017direct,newcombe2011dtam}. However, the direct method is based on the assumption that the gray level is invariant. This assumption may be invalid in large scale scene.  
In addition to monocular SLAM, some stereo SLAM systems have been proposed. ORB-SLAM2 (stereo) \cite{mur2017orb}, proSLAM \cite{schlegel2018proslam} and Vins-Fusion (stereo)  can recover the scale of localization and point cloud relying on the camera extrinsic parameters. However, the scale recovered by extrinsic parameters of cameras is inaccurate in large-scale scene. Thus, GNSS information is used to assist localization in our work.

Global Navigation Satellite System (GNSS) can provide relatively accurate positioning in large-scale scenes, and it is observable for scale. Therefore, visual-odometry (VO) system can be coupled with GNSS for optimizing global positioning, relative positioning and restoration of scale. Qin \cite{qin2019general} propose a framework for coupling global sensors (GNSS, barometer, magnetometer, etc) with local odometry.  This framework can accommodate a variety of sensors, but only when the displacement is sufficient, the scale can be accurately recovered. Simon \textit{et al.} \cite{boche2022visual} propose a VI-GPS tightly coupled SLAM system that achieves state-of-the-art performance. However, inertial measurement units may not be effective for large-scale scenarios.

\subsection{Dense Reconstruction}

The dense reconstruction techniques based on visual geometric features have been widely used in depth cameras, offline map reconstruction, etc. The SfM method has been quite comprehensive in the past, but it is limited by the algorithm framework and cannot achieve real-time 3D reconstruction. 

Stereo-matching algorithms are the key components for 3D reconstruction, which find corresponding pixels from multiple images with overlapping perspectives and estimate the dense point cloud map of a 3D scene. Traditional stereo-matching algorithms can be divided into local matching methods and global energy minimization methods. For large-scale images, local matching methods usually behaves more efficiently due to the restricted search domain. The method proposed by Bleyer \textit{et al.} \cite{bleyer2011patchmatch} applys slanted support windows and computes 3D planes at the pixel level. The local stereo matching is performed in near-constant time with ELAS \cite{geiger2010efficient}. The algorithm generates a 2D mesh using robust support points from rectified stereo images via Delaunay triangulation. Bayesian inference is used to match pixels inside each triangle quickly and independently based on their disparities.

The field of multi-view reconstruction based on deep learning has developed rapidly in the past few years. Yao \textit{et al.} \cite{yao2018mvsnet} propose a standard end-to-end multi-view stereo depth estimation network, which can recover the depth of an image and in turn can accomplish dense reconstruction of point clouds.
A method called Soft Argmin \cite{kendall2017end} is used to fit the depth, which can solve the problem to some extent. Since the depth map is predicted on a per-view basis, errors in the predictions will lead to inconsistencies when combining multiple local reconstructions into the global 3D model. 
NeRF \cite {mildenhall2021nerf} is proposed by Mildenhall, Ben \textit{et al.}, which is a implicit scene representation. There has been a lot of work on scene reconstruction based on NeRF with good results, for example Manhattan-SDF \cite{guo2022neural}. However, these methods are difficult to be applied for real-time reconstruction in large-scale scenes.

\section{Methodology}

The framework of cluster reconstruction system is depicted in Fig. \ref{fig:pipeline} and it primarily consists of four components: ego-pose estimation, dense reconstruction, relative pose estimation, and point cloud fusion. The pipeline of framework will be introduced in the following subsections. 
To save space, we only present the residual formulas for the optimization problem in the later method description. The complete optimization equation is shown as below,

\begin{equation}
	\min _ {\boldsymbol{\mathcal{X}}} \left\{\sum_{n=1}^ {\boldsymbol{\mathbf{\mathcal{K}}}} \boldsymbol{e}_n^{T}(\boldsymbol{\mathcal{X}}) \boldsymbol{W}_n \boldsymbol{e}_n(\boldsymbol{\mathcal{X}})\right\},
\end{equation}
where $\boldsymbol{\mathcal{X}}$ is the variable to be optimized; $\boldsymbol{\mathcal{K}}$ is the set of constructed residuals; $\boldsymbol{e}_n$ is the residual and $\boldsymbol{W}_n$ is the information matrix, which represents the weights of the residuals.

\subsection{Ego-pose and Relative-pose Estimation} 

\begin{figure}[tb]
	\centering
	\includegraphics[width=1\linewidth]{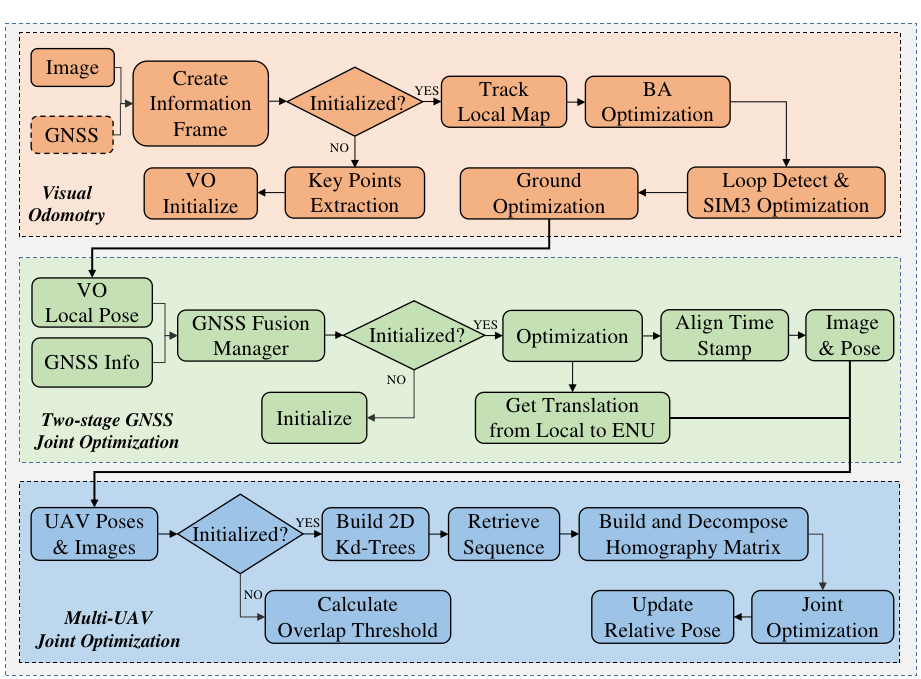}
	\caption{The multi-UAV localization system is divided into three parts: VO, two-stage GNSS joint optimization and multi-UAV joint optimization. The VO component uses the large-scale images collected to estimate the ego-pose. The two-stage GNSS joint optimization component optimizes the global poses of UAVs and calculates their initial relative poses using GNSS information. The multi-UAV joint optimization component utilizes the initial relative pose information to identify overlapping images and improves the relative pose through the common-view relationship between images.}

	
	\label{fig:localization_pipline}
\end{figure}

Although significant progress has been made in single-UAV state estimation, multi-UAV cooperative localization for cluster mapping still faces challenges due to limited communication bandwidth, computational power, positioning accuracy, and scale drift. 
To address these issues, a lightweight and scalable multi-UAV localization system utilizing GNSS information and co-viewing relationships has been implemented to satisfy the needs for relative pose estimation and scale recovery.
The single-UAV localization consists of a vision based odometry front-end and a two-stage joint optimization back-end. 
Once individual UAV poses have been obtained, the overlapping observation areas of multiple UAVs are automatically identified and a relative pose optimization method is applied to optimize their relative poses. Our proposed method fully utilizes the prior information provided by GNSS to accelerate the retrieval of overlapping regions.
To maximize the use of multi-core CPUs and enhance the real-time performance of the system, this system employs multiple threads for parallel processing. The flowchart of the multi-UAV localization system is depicted in Fig. \ref{fig:localization_pipline}.

\subsubsection{Visual Odometry}

Accurate and efficient pose estimation is crucial for 3D reconstruction of large-scale scenes. However, the ORB feature point extraction and matching method has limitations in terms of accuracy and computational efficiency when dealing with high-resolution images captured by UAVs. We adopt SiftGPU \cite{wu2011siftgpu} for feature point detection and descriptor calculation. This method not only offers high robustness but also takes advantage of the parallel computing capabilities of the GPU to achieve real-time processing speeds. Once the matching feature points between two frames have been obtained, an optimization problem is formulated to determine the UAV poses.

Considering that the optimization function can be adapted to different camera models, the matching points in this section are assumed to reside on the normalized plane. Let $\boldsymbol{p}_{c_i}\left(u_{c_i}, v_{c_i}\right)$ and $\boldsymbol{p}_{c_j}\left(u_{c_j}, v_{c_j}\right)$ be the same 3D point observed in two frames on the normalized plane. According to the re-projection formula, the $\boldsymbol{p}_{c_i}$ point can be re-projected to the camera normalization plane of the frame where $\boldsymbol{p}_{c_j}$ is located to construct the residual function.  The three-dimensional coordinate is denoted as $\boldsymbol{P}_{c_j}\left(X_{c_j}, Y_{c_j}, Z_{c_j}\right)$ which is recovered from the normalized coordinate of $\boldsymbol{p}_{c_i}\left(u_{c_i}, v_{c_i}\right)$ re-projected from the $i$ frame to the $j$ frame. The reprojection formula can be written as follows:
\begin{equation}
	\left[\begin{array}{c}
		\boldsymbol{P}_{c_j} \\
		1
	\end{array}\right]= \boldsymbol{T}_{w c_j}^{-1} \boldsymbol{T}_{w c_i}  \frac{1}{\lambda}\left[\begin{array}{c}
		u_{c_i} \\
		v_{c_i} \\
		1 \\
		1
	\end{array}\right],
\end{equation}
where $\boldsymbol{T}_{w c_i}$ is the transformation from the camera coordinate system of the $i^{th}$ frame to the initial world coordinate system, which means the pose of the $i^{th}$ frame in the world coordinate system. $\boldsymbol{T}_{w c_j}^{-1}$ is the transformation from the world coordinate system of the $j^{th}$ frame to the camera coordinate system. $\lambda$ is the depth of feature points. Visual residual can finally be expressed as:
\begin{equation}
	\boldsymbol{e}_r^{i, j}=\left[\begin{array}{l}
		\frac{X_{c j}}{Z_{c j}}-u_{c j} \\
		\frac{Y_{c j}}{Z_{c j}}-v_{c j} 
	\end{array}\right].
\end{equation}

After obtaining the relative poses between images through the optimization, the depth information of the sparse point cloud is recovered by triangulation. This depth information is then used for dense reconstruction. 
If GNSS information is available, the image poses and the sparse point cloud are transformed to the Earth-Centered, Earth-Fixed (ECEF) coordinate system to obtain a global position for the reconstructed area.

\subsubsection{Two-stage Joint Optimization}

A two-stage joint optimization algorithm is applied to estimate the relative pose of multiple UAVs and to calculate the spatial scale of point clouds and poses. In the VO front-end, the local poses $\boldsymbol{T}_{w c}$ of the UAVs are obtained. 
%
In the first stage, these local poses are then combined with observation information obtained from GNSS to recover the scale of the relative poses and optimize the current global poses.
In the second stage, the relative poses of multiple UAVs are estimated using GNSS data.
The first frame of GNSS data is used to establish the initial ENU (East-North-Up) coordinate system when performing initialization, and subsequent GNSS data will be transformed to it.
The VO initial local coordinate system is set as the origin world coordinate at the same time. 

After obtaining both VO and GNSS data, the timestamps of two sources are aligned, and an optimization function is constructed to estimate the global pose and compute the metric scale. Before the optimization process, an initial value for the scale parameter $s$ must be solved, 
\begin{equation}
	s=\frac{\sqrt{g_{x_k}^2+g_{y_k}^2+g_{z_k}^2}}{\sqrt{\left(p_{x_k}-p_{x_0}\right)^2+\left(p_{y_k}-p_{y_0}\right)^2+\left(p_{z_k}-p_{z_0}\right)^2}},
\end{equation}
where  $(g_{x_k}, g_{y_k}, g_{z_k})$ represents the translation from the current ENU coordinate system to the initial ENU coordinate system. $(p_{x_k}, p_{y_k}, p_{z_k})$ represents the translation of the current frame in the VO origin coordinate system, while $(p_{x_0}, p_{y_0}, p_{z_0})$ is the initial translation in the VO origin coordinate system.
The two-stage joint optimization residual function consists of two parts. The first part can be expressed as:

\begin{equation}
	\left\{\begin{array}{l}
		\boldsymbol{e}_q=\left(\boldsymbol{q}_j^l \boldsymbol{q}_i^{l^{-1}}\right) \boxminus\left(\boldsymbol{q}_j^w \boldsymbol{q}_i^{w^{-1}}\right), \\
		\boldsymbol{e}_t=s \boldsymbol{q}_i^{l^{-1}}\left(\boldsymbol{t}_j^l-\boldsymbol{t}_i^l\right)-\boldsymbol{q}_i^{w^{-1}}\left(\boldsymbol{t}_j^w-\boldsymbol{t}_i^w\right),
	\end{array}\right.
\end{equation}
where $\boldsymbol{e}_q$ is the rotation constraint generated by the estimated pose of VO. $\boldsymbol{e}_t$ is the translation constraint generated by the pose estimated by VO. $\boldsymbol{q}_j^l$, $\boldsymbol{q}_i^l$, $\boldsymbol{t}_j^l$, $\boldsymbol{t}_i^l$ represent the pose transformation between two adjacent coordinates after aligning with the GNSS timestamps in the VO local coordinate system respectively. $s$ is the scale to be optimized. $\boldsymbol{q}_j^w$, $\boldsymbol{q}_i^w$, $\boldsymbol{t}_j^w$, $\boldsymbol{t}_i^w$ represent the pose transformation between two adjacent coordinates in the ENU coordinate system, which are also the variables to be optimized respectively. $\boxminus$ is an operation on the quaternion error state. 
The second part is the constraints imposed by GNSS observation information: 
\begin{equation}
	\left\{\begin{array}{l}
		\boldsymbol{e}_g=\boldsymbol{t}_k^{GNSS}-\boldsymbol{t}_k^w, \\
		\boldsymbol{e}_s=s-\frac{\bmod \left(\boldsymbol{t}_k^{GNSS}\right)}{\bmod \left(\boldsymbol{t}_k^w\right)},
	\end{array}\right.
\end{equation}
where $\boldsymbol{t}_k^{GNSS}$ is the translation of the current frame to the GNSS initial coordinate system. $\boldsymbol{t}_k^w$ is the translation of the current frame in the world coordinate system. $s$ is the scale to be optimized. After evaluation, the optimization equation constructed above can converge quickly and calculate the scale of the pose accurately. Since the calculation of the dense point cloud map requires the pose of each image in the camera coordinate system, it is necessary to transform the optimized pose into the camera coordinate system.

In addition to recovering the scale of the camera poses and sparse point cloud, the two-stage localization algorithm can also estimate the relative poses. 
To estimate the relative poses of multiple UAVs, we define a central UAV whose pose is represented as $\boldsymbol{R}_{lw}^0, \boldsymbol{P}_{GNSS}^0$. Here, $\boldsymbol{R}_{lw}^0$ denotes the rotation of the central VO local coordinate system to the world coordinate system (ENU). Similarly, $\boldsymbol{P}_{GNSS}^0$ represents the longitude, latitude, and altitude of the central UAV's GNSS original coordinate system (ENU).
For the $k^{th}$ UAV, its pose is defined as $\boldsymbol{R}_{lw}^k, \boldsymbol{P}_{GNSS}^k$. As a result, the rotation of the  $k^{th}$ UAV to the central UAV is $\boldsymbol{R}_{lw}^{0^{-1}}\boldsymbol{R}_{lw}^k$, and the translation is $\boldsymbol{t}_{0k}$ which is calculated by  $\boldsymbol{P}_{GNSS}^0,\boldsymbol{P}_{GNSS}^k$. $\boldsymbol{T}_{0k}$ is used to represent the transformation from the initial coordinate system of the $k^{th}$ UAV to the initial coordinate system of the center UAV for the convenience of later expression. 

In large-scale and long-term reconstruction, the two-stage joint GNSS localization algorithm can effectively reduce the positioning drift. According to $\boldsymbol{T}_k = \boldsymbol{T}_{lw}^{0^{-1}}\boldsymbol{T}_{0k}\boldsymbol{T}_{lw}^k$, the relative poses can be updated in real-time. 

\subsubsection{Relative Pose Optimization}
While the algorithm described above enables the determination of the initial relative pose between multiple UAVs, relying solely on GNSS information for point cloud alignment may not yield the desired level of accuracy due to errors in GNSS data.
To improve the accuracy of relative poses between multiple UAVs, it is necessary to incorporate additional information from images for further optimization. 
The focus of the framework proposed in this paper is dense reconstruction in large-scale scenes, hence the acquired images conform well to the planar hypothesis and can be used to achieve accurate relative positioning of UAVs.

We propose a method that uses the homography matrix to optimize the relative positions of multiple UAVs. Due to computational limitations, the relative poses of each pair of UAVs are optimized one by one. 
First, we utilize the relative pose obtained from the previously mentioned two-stage joint optimization algorithm to select candidate overlapping image pairs. 
The desired distance is calculated using the overlap threshold, camera intrinsic parameters, and flight altitude. 
We construct KD-trees using the image position sequences of UAVs respectively to accelerate the retrieval, and retrieve the sequence that matches the desired distance. Homography matrices are constructed using corresponding image pairs from different sequences. By decomposing the matrix, positional observations can be obtained between the image sequences.

The observation pose based on homography decomposition is denoted as $\boldsymbol{q}_{c_1 c_2}^H, \boldsymbol{t}_{c_1 c_2}^H$. The relative position between two UAVs can be found by calculating the transformation from each UAV to the central UAV. The transformation is defined as: 
\begin{equation}
	\begin{aligned}
		& \boldsymbol{q}_{c_1 c_2} = \boldsymbol{q}_{\text {core}\_c_1} \boldsymbol{q}_{k_1}^{c_1}\left(\boldsymbol{q}_{\text {core}\_c_2}\boldsymbol{q}_{k_2}^{c_2}\right)^{-1}, \\
		& \boldsymbol{t}_{c_1 c_2} = \boldsymbol{t}_{k_1}^{c_1}-\boldsymbol{q}_{\text {core}\_c_1}^{-1}\left(\boldsymbol{q}_{\text {core}\_c_2} \boldsymbol{t}_{k_2}^{c_2}+t_{\text {core}\_c_2}-\boldsymbol{t}_{\text {core}\_c_1}\right),
	\end{aligned}
\end{equation}
where $\boldsymbol{q}_{\text {core}\_c_1}, \boldsymbol{t}_{\text {core}\_c_1} $ and $\boldsymbol{q}_{\text {core}\_c_2}, \boldsymbol{t}_{\text {core}\_c_2}$ represent the transformations from the coordinate systems of the two UAVs to the coordinate system of the central UAV; $\boldsymbol{q}_{k_1}^{c_1}, \boldsymbol{t}_{k_1}^{c_1}$ and $\boldsymbol{q}_{k_2}^{c_2}, \boldsymbol{t}_{k_2}^{c_2}$ represent the pose of two UAVs. Before optimizing the relative poses, it is necessary to align the scales between the relative poses of the UAV and the transformations derived from the homography matrix decomposition. The scale can be recovered by: 
\begin{equation}
	\boldsymbol{t}_{c_1 c_2}^{H_s}=\frac{\left\|\boldsymbol{t}_{c_1 c_2}\right\|_2}{\left\|\boldsymbol{t}_{c_1 c_2}^H \right\|_2} \cdot \boldsymbol{t}_{c_1 c_2}^H. 
\end{equation}
The residual formula is constructed as follows:
\begin{equation}
	\left\{\begin{array}{l}
		\boldsymbol{e}_q=\boldsymbol{q}_{c_1 c_2} \boxminus \boldsymbol{q}_{c_1 c_2}^H, \\
		\boldsymbol{e}_t=\boldsymbol{t}_{c_1 c_2}-\boldsymbol{t}_{c_1 c_2}^{H_s}.
	\end{array}\right.
\end{equation}
The variables to be optimized are  $\boldsymbol{q}_{\text {core}\_c_1}, \boldsymbol{t}_{\text {core}\_c_1} $ and $\boldsymbol{q}_{\text {core}\_c_2}, \boldsymbol{t}_{\text {core}\_c_2} $. Image information from multiple UAVs is used to improve the accuracy of relative poses, which will be used in the later point cloud fusion stage.

\subsection{Mapping}

The mapping process involves the reconstruction of a dense 3D map and can be divided into two main stages: densification and fusion. Densification involves the reconstruction of a 3D map by a single UAV, while the map tiles reconstructed by multiple UAVs are fused together in the second stage, called fusion.

\subsubsection{Densification}

After obtaining accurate relative pose estimations between image frames through the co-localization process, the dense mapping process can be performed. 
In small-scale or indoor reconstruction scenes, a stereo camera is commonly used to generate a dense point cloud. However, in large-scale scenes, the limited baseline of stereo cameras makes it challenging to generate accurate dense point clouds in real-time. We construct virtual stereo pairs using the local accurate pose estimated by VO and images captured at two different moments. This allows for both accurate calculations and real-time performance. Dense point cloud maps are generated using depth maps. To improve the matching efficiency, we use the method proposed by Bouguet \cite{bouguet2008camera} to maximize the common area of the left and right views. In the process of constructing dense disparity maps, the fast stereo matching method ELAS \cite{geiger2010efficient} is employed.

\begin{figure*}[]
	\centering
	\subfigure[\textit{ClusterFusion}]{
		\includegraphics[width=0.8\textwidth]{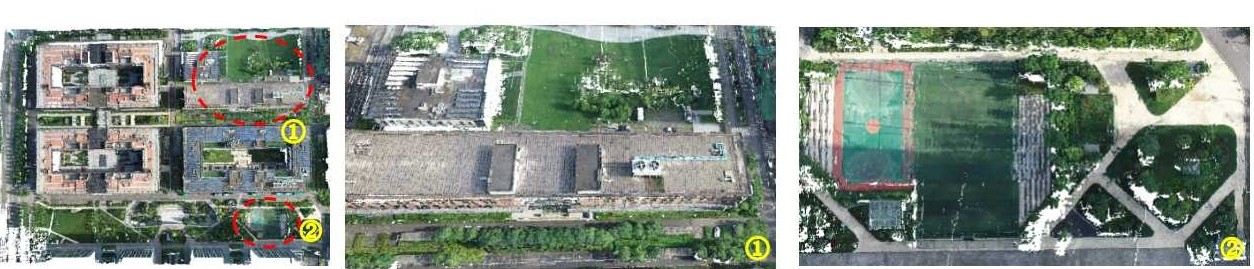}
		\label{fig:ClusterFusion_details}
	}
	\subfigure[\textit{DenseFusion}]{
		\includegraphics[width=0.8\textwidth]{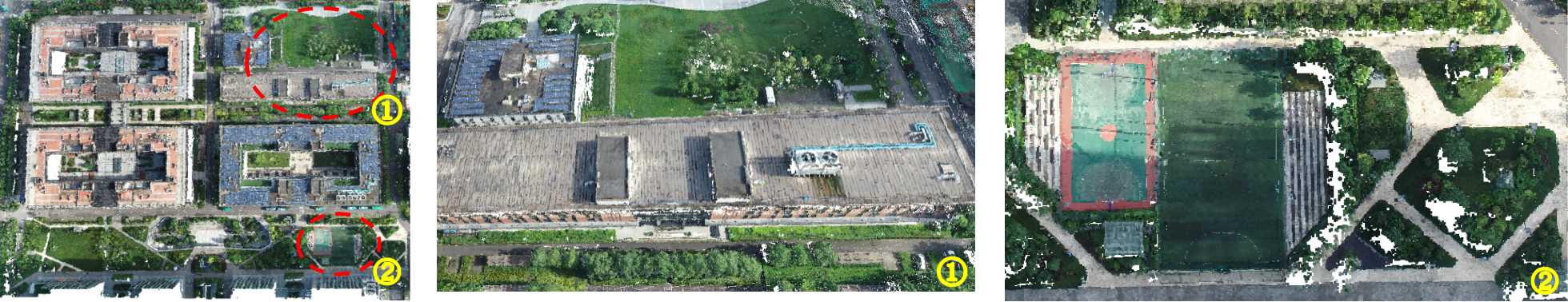}
		\label{fig:DenseFusion_details}
	}
	\caption{This set of images shows the details of the dense reconstruction using the \textit{Huanpu} dataset. The first row of images displays the details of the dense point cloud generated by ClusterFusion, while the second row displays the images generated by DenseFusion.}
	\label{fig:details of reconstructure}
\end{figure*}
\begin{figure*}[]
	\centering
	\subfigure[\textit{ClusterFusion}]{
		\includegraphics[width=0.8\textwidth]{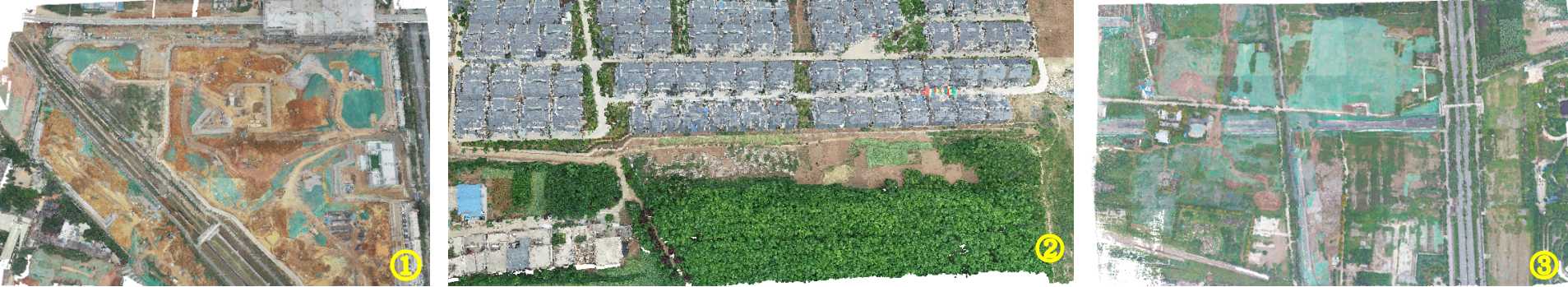}
		\label{fig:ClusterFusion_large}
	}
	\subfigure[\textit{DenseFusion}]{
		\includegraphics[width=0.8\textwidth]{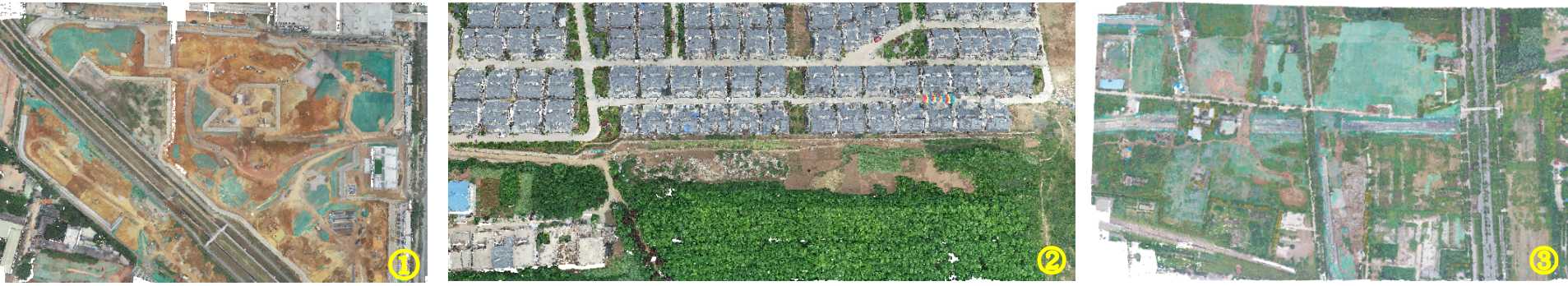}
		\label{fig:DenseFusion_large}
	}
	\caption{This set of images shows the effect of dense reconstruction in large-scale scene. The images in these two rows represent the reconstruction results using ClusterFusion and DenseFusion, respectively. Dataset 1 is \textit{Factory}, dataset 2 is \textit{Wutai} and dataset 3 is \textit{Yuhuazhai}.}
	\label{fig:reconstruction_of_large}
\end{figure*}

\subsubsection{Fusion}

In the point cloud fusion process, the central UAV receives the dense 3D point cloud map constructed by each UAV, as well as the real-time relative pose between UAVs. For a point $\boldsymbol{p}_t^{k}$ in the dense point cloud reconstructed by the $k^{th}$ UAV, it is transformed into the point cloud map constructed by the central UAV as follows: $\boldsymbol{p}_t^{0} = \boldsymbol{T}_k ^{-1}\boldsymbol{p}_t^{k}$. The fused dense point cloud map may have overlap due to overlapping scenes, which significantly increases the memory consumption and computational complexity of point cloud processing. To eliminate this overlap,  voxelization method is used to downsample the point cloud map, which can reduce the scale of the point cloud to a certain extent. As our method is suitable for large-scale scenes, we use location information to segment the point cloud and store it in different point clouds to facilitate writing the inactive part of the point cloud, thereby saving memory. The index of each point is calculated as follows:
\begin{equation}
	\begin{aligned}
		&Ind_x=[P_x / { d_c }], \\
		&Ind_y=[P_y / { d_c }],
	\end{aligned}
\end{equation}
where $P_x, P_y $ are the coordinates of the point cloud to the position on the $x-y$ plane in the central coordinate system. $d_c$ is the width of the point cloud block. Storing point cloud information in this way can not only save memory, but also effectively limit the scale of point clouds in operations such as voxelization of back-end point clouds.

\section{Experiments}

We implement a real-time multi-UAV collaborative mapping framework using C++ and evaluate its performance using a dataset of multi-UAV mapping in large-scale scenes. The dataset is mainly obtained through aerial photography in real-world. The dataset comprises high-resolution image data obtained by multiple UAVs, along with corresponding GNSS data. Each image contains over 12,000,000 pixels and covers various scenes, including urban and farmland areas. The dataset is suitable for applications such as map production, collaborative mapping using multiple UAVs, and machine learning algorithm training. The point cloud map constructed using this dataset is shown in Fig. \ref{fig:details of reconstructure} and \ref{fig:reconstruction_of_large}. It can be seen that the point cloud map constructed by multi-UAV collaborative mapping is similar to DenseFusion in accuracy. 
In the \textit{Huanpu} dataset, both methods are effective in restoring the details of the buildings and playgrounds. When it comes to large-scale scenes, the point cloud maps produced by the approach presented in this paper are comparable to DenseFusion, especially in the \textit{Yuhuazhai} scene which conforms better to the planar hypothesis.

In order to verify the performance of the proposed framework, we design three experiments to compare the robustness, speed and accuracy of localization and mapping. Experiments are performed on a desktop computer with  Intel(R) Xeon(R) CPU E5-2676 v3 @ 2.40GHz, GeForce GTX 1080 and 32G RAM. 
%
Firstly, the scale recovery speed and stability of two algorithms are compared. Secondly, the reconstruction speed of our method is compared with the previous research DenseFusion \cite{chen2020densefusion}. Finally, the reconstruction accuracy are evaluated by comparing our algorithm with SIBITU \footnote{www.sibitu.cn}, and the results demonstrate that our method achieves similar reconstruction accuracy.

\subsection{GNSS optimization}

For dense reconstruction of UAV clusters, aligning the pose estimation scales between UAVs is crucial for successful point cloud fusion. 

\begin{figure}[tb]
	\centering
	\includegraphics[width=0.8\linewidth]{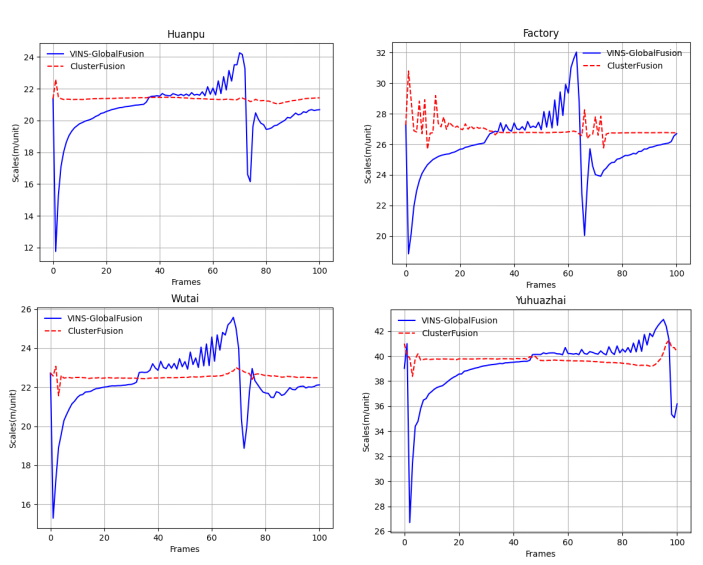}
	\caption{Scale results from VINS-GlobalFusion and ClusterFusion. }
	\label{fig:scale}
\end{figure}

The translation scale for a single UAV is calculated, and then compared with the scale calculated using method VINS-GlobalFusion \cite{qin2019general}. The results are shown in Fig. \ref{fig:scale}. The experimental results demonstrate that the proposed method can calculate the scale of pose estimation faster and more stably. As our data analysis suggests limited scale drift in the visual odometry, the estimated scale should fluctuate with GNSS errors and remain relatively stable. Comparison of data generated from multiple datasets shows that our proposed method effectively eliminates GNSS data errors and improves the robustness of scale estimation.

\begin{table}[tb]
	\caption{The running speed of DenseFusion and ClusterFusion.}
	\label{table:time_statistics}
	\centering
	\begin{tabular}{|l|c|c|c|c|c|}
		\hline
		Sequence              & Images & Resolution  & DenseFusion (s)  & ClusterFusion (s)     \\ \hline
		\textit{Factory}      & 365    & 4864*3648   &   331        &   \textbf{188}       \\
		\textit{Huanpu}       & 210    & 5472*3648   &   145        &   \textbf{81}        \\
		\textit{Wutai}        & 318    & 5472*3648   &   331        &   \textbf{206}       \\
		\textit{Yuhuazhai}    & 165    & 5472*3648   &   292        &   \textbf{81}        \\
		\textit{Famensi}      & 381    & 4000*3000   &   436        &   \textbf{130}       \\ \hline
	\end{tabular}
\end{table}

\begin{figure}[tb]
	\centering
	\subfigure[\textit{Huanpu}]{
		\includegraphics[width=0.4\textwidth]{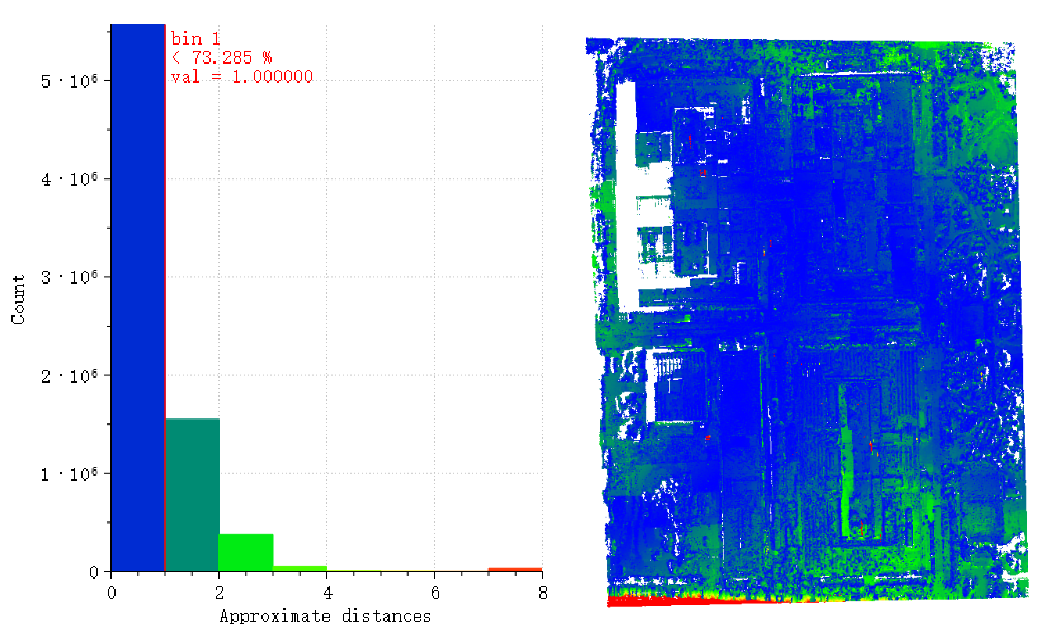}
		\label{fig:accuracy_huanpu}
	}
	\subfigure[\textit{Yuhuazhai}]{
		\includegraphics[width=0.4\textwidth]{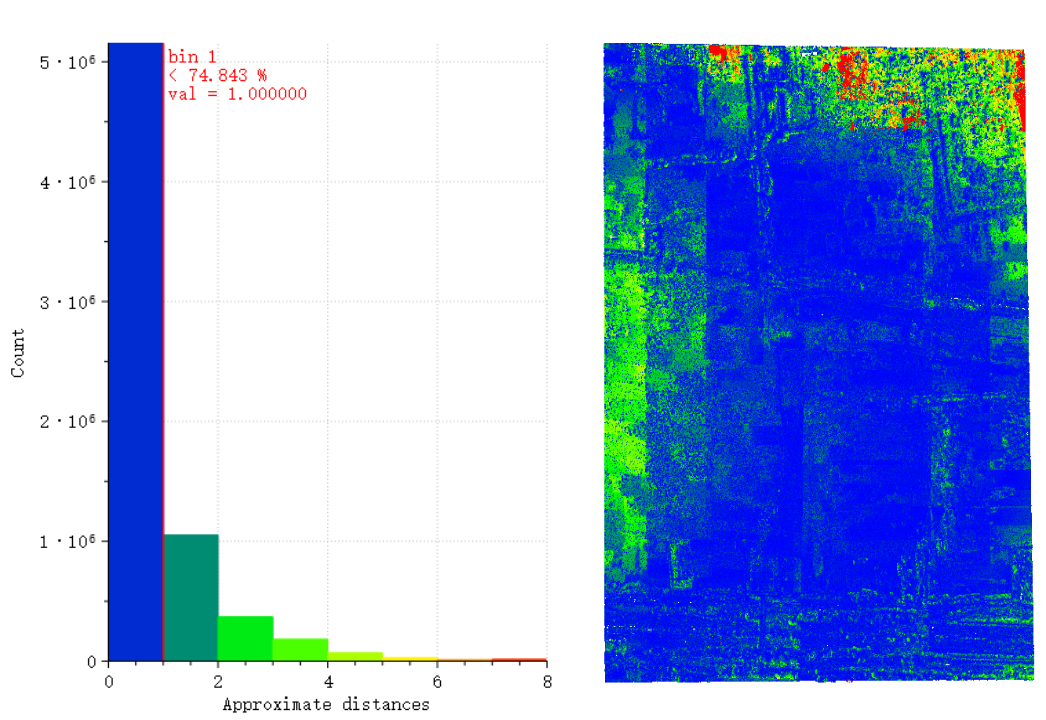}
		\label{fig:accuracy_yuhuazhai}
	}
	\caption{Accuracy evaluation results of the reconstruction.}
	\label{fig:accuracy}
\end{figure}

\subsection{Computational Performance}

The performance is evaluated by comparing the proposed multi-UAV collaborative reconstruction framework with Densfusion, using the same equipment and datasets. The comparison results are shown in Table \ref{table:time_statistics}. 
The results show that the efficiency of ClusterFusion is higher than DenseFusion under the same construction area. Thanks to the well-designed architecture for multi-UAV reconstruction, multiple UAVs can collaborate on positioning, reconstruction, and point cloud fusion without consuming too much computing resources and bandwidth. This greatly improves reconstruction efficiency. 
Running this experiment on a single computation platform limits the potential speedup. If we distribute the framework across multiple computing platforms, the computation speed can be further improved.

The computation speed of 3D reconstruction is impacted not only by the number of UAVs but also by the resolution of images. This impact is particularly noticeable in the stage of achieving 3D through matched pixel points from virtual stereo pairs. To improve the efficiency of the algorithm, utilizing the parallel computing power of the GPU can be a valuable optimization strategy.

\subsection{Accuracy Evaluation}

For the accuracy evaluation of point clouds, there is currently no perfect evaluation method. The point cloud constructed by the commercial software SIBITU based on the SfM method has theoretically higher accuracy. We propose using our framework to generate point cloud maps for various scenarios in the dataset, and then comparing them to the offline point cloud maps constructed by SIBITU software. We use the least squares plane method to calculate the distance between the two point clouds. This approach enables us to evaluate the effectiveness of our framework in generating accurate point cloud maps for different scenarios. Some evaluation results are shown in Fig. \ref{fig:accuracy}. The results show that in dataset \textit{Huanpu}, the errors of 93.7\% points are less than 2.0 m, and 74.2\% are less than 1.0 m; in dataset \textit{Yuhuazhai}, the errors of 90.1\% points are less than 2.0 m and 74.8\% are less than 1.0 m.
The accuracy of mapping is mainly affected by the pose estimation of a single UAV and the relative pose estimation accuracy between multiple UAVs. Through the bilateral filtering algorithm, certain bad points can be removed and the quality of the point cloud can be improved. Compared with SIBITU, the accuracy of this framework is relatively close, and the speed is several times faster.

\section{Conclusions}
In this paper, we present a collaborative real-time dense reconstruction framework for multiple UAVs which includes visual odometry, two-stage joint GNSS optimization, densification, relative pose optimization, and dense point cloud map fusion. 
Compared to previous dense reconstruction frameworks, the proposed framework offers several advantages. 
%
Firstly, it is faster than previous work, with the same arithmetic platform, processing at more than twice the speed. Secondly, the framework can be ported to cluster UAVs for joint reconstruction. The computational efficiency and robustness will be further improved on a distributed platform. Finally, it is scalable, supporting expansion of the number of UAVs with low cost.

The proposed framework performs better in large-scale dense reconstruction tasks, but there are still aspects to consider in future research. Enhancing adaptability of the framework to different environments would be beneficial. Optimizing the framework and algorithm to fully utilize the parallel computing performance of the GPU could improve performance. Lastly, real-time map analysis processing is necessary after completing dense reconstruction to enable rapid response to changes in the environment. 




\bibliographystyle{./IEEEtranBST/IEEEtran}
\bibliography{./ClusterFusion}

\end{document}